## *Engineering Note*

# Optiplan: Unifying IP-based and Graph-based Planning

**Menkes H.L. van den Briel**                                         MENKES@ASU.EDU
*Department of Industrial Engineering*
*Arizona State University, Tempe, AZ 85281 USA*

**Subbarao Kambhampati**                                         RAO@ASU.EDU
*Department of Computer Science and Engineering*
*Arizona State University, Tempe, AZ 85281 USA*

## Abstract

The Optiplan planning system is the first integer programming-based planner that successfully participated in the international planning competition. This engineering note describes the architecture of Optiplan and provides the integer programming formulation that enabled it to perform reasonably well in the competition. We also touch upon some recent developments that make integer programming encodings significantly more competitive.

## 1. Introduction

Optiplan is a planning system that uses integer linear programming (IP) to solve STRIPS planning problems. It is the first such system to take part in the international planning competition (IPC) and was judged the second best performer in four competition domains of the optimal track for propositional domains. Optiplan's underlying integer programming formulation extends the state change model by Vossen and his colleagues (1999). Its architecture is very similar to that of Blackbox (Kautz & Selman, 1999) and GP-CSP (Do & Kambhampati, 2001), but instead of unifying satisfiability (SAT) or constraint satisfaction (CSP) with graph based planning, Optiplan uses integer programming. Like Blackbox and GP-CSP, Optiplan works in two phases. In the first phase a planning graph is built and transformed into an IP formulation, then in the second phase the IP formulation is solved using the commercial solver ILOG CPLEX (ILOG Inc., 2002).

A practical difference between the original state change model and Optiplan is that the former takes as input all ground actions and fluents over all initialized plan steps, while the latter takes as input just those actions and fluents that are instantiated by Graphplan (Blum & Furst, 1995). It is well known that the use of planning graphs has a significant effect on the size of the final encoding no matter what combinatorial transformation method (IP, SAT, or CSP) is used. For instance, Kautz and Selman (1999) as well as Kambhampati (1997) pointed out that Blackbox's success over Satplan (Kautz & Selman, 1992) was mainly explained by Graphplan's ability to produce better, more refined, propositional structures than Satplan. In addition, Optiplan allows propositions to be deleted without being required as preconditions. Such state changes are not modeled in the original state change model, and therefore Optiplan can be considered to be a more general encoding. One more, although minor, implementation detail between Optiplan and the state change model is that Optiplan reads in PDDL files.

This engineering note is organized as follows. Section 2 provides a brief background on integer programming and Section 3 discusses previous IP approaches to planning. Section 4 describes the Optiplan planning system and its underlying IP formulation. In Section 5 we give some experimental results and look at Optiplan's performance in the international planning competition of 2004 (IPC4). Conclusions and a brief discussion on some recent developments is given in Section 6.

## 2. Background

A linear program is represented by a linear objective function and a set of inequalities, such as $\min\{cx : Ax \geq b, x \geq 0\}$ where $x$ an $n$-dimensional column vector of variables, A is an $m$ by $n$ matrix, $c$ an $n$-dimensional row vector, and $b$ an $m$-dimensional column vector. If all variables are constrained to be integers then we have an integer (linear) program, and if all variables are restricted to 0-1 values then we have a binary integer program.

The most widely used method for solving general integer programs is by using branch and bound on the linear programming relaxation. Branch and bound is a general search method in which subproblems are created that restrict the range of the integer variables, and the linear programming relaxation is a linear program obtained from the original integer program by omitting the integrality constraints. An *ideal* formulation of an integer program is one for which the solution of the linear programming relaxation is integral. Even though every integer program has an ideal formulation (Wolsey, 1998), in practice it is very hard to characterize the ideal formulation as it may require an exponential number of inequalities. In problems where the ideal formulation cannot be determined, it is often desirable to find a *strong* formulation of the integer program. Suppose that $P_1 = \min\{cx : A_1 x \geq b_1, x \geq 0\}$ and $P_2 = \min\{cx : A_2 x \geq b_2, x \geq 0\}$ are the linear programming relaxations of two IP formulations of a problem, then we say that formulation $P_1$ is stronger than formulation $P_2$ if $P_1 \subset P_2$. That is, the set of solutions of $P_1$ is subsumed by the set of solutions of $P_2$.

## 3. Integer Programming Approaches to Planning

Despite the vast amount of research that has been conducted in the field of AI planning, the use of linear programming (LP) and integer linear programming have only been explored at a marginal level. This is quite surprising since (mixed) integer linear programming provide feasible environments for using numeric constraints and arbitrary linear objective functions, two important aspects in real-world planning problems.

Only a handful of works have explored the use of LP and IP techniques in AI planning. Bylander (1997) developed an IP formulation for classical planning and used the LP relaxation as a heuristic for partial order planning. The results, however, do not seem to scale well compared to planning graph and satisfiability based planners.

The difficulty in developing strong IP formulations is that the performance often depends on the way the IP formulation is constructed. Vossen et al. (1999) compared two formulations for classical planning. First, they consider a straightforward IP formulation based on converting the propositional representation given by Satplan (Kautz & Selman, 1992) to an IP formulation with variables that take the value 1 if a certain proposition is true, and 0 otherwise. In this formulation, the assertions expressed by IP constraints directly correspond to the logical conditions of the propositional representation. Second, they consider an IP formulation in which the original propositional variables are replaced by *state change* variables. State change variables take the value 1 if a certain proposition is added, deleted, or persisted, and 0 otherwise. Vossen et al. show that the formulation based on state change variables outperforms the straightforward formulation based on converting the propositional representation.

Dimopoulos (2001) improves the IP formulation based on state change variables by identifying valid inequalities that tighten the formulation. Yet, even stronger IP formulations are given by Bockmayr and Dimopoulos (1998, 1999), but their IP formulations contain domain dependent knowledge and are, therefore, limited to solving problems of specific problem domains only.

LP and IP techniques have also been explored for non-classical planning. Dimopoulos and Gerevini (2002) describe a mixed integer programming formulation for temporal planning and Wolfman and Weld (1999) use an LP formulation in combination with a SAT formulation to solve resource planning problems. Kautz and Walser (1999) also use IP formulations for resource planning problems but, in addition, incorporate action costs and complex objectives.

So far, none of these IP approaches to AI planning ever participated in the IPC, making it harder to assess the relative effectiveness of this line of work. Optiplan, a planner based on the state change formulation, is the first IP-based planner to do so.

## 4. Optiplan

Optiplan is a planning graph based planner and works as follows. First we build the planning graph to the level where all the goal fluents appear non-mutex. Then we compile the planning graph into an integer program and solve it. If no plan is found, the planning graph is extended by one level and the new graph is again compiled into an integer program and solved again. This process is repeated until a plan is found.

In the remainder of this section we give the IP formulation that is used by Optiplan. In order to present the IP formulation we will use the following notation. $\mathcal{F}$ is the set of fluents and $\mathcal{A}$ is the set of actions (operators). The fluents that are true in the initial state and the fluents that must be true in the goal are given by $\mathcal{I}$ and $\mathcal{G}$ respectively. Furthermore, we will use the sets:

- $pre_f \subseteq \mathcal{A}$, $\forall f \in \mathcal{F}$, set of actions that have fluent $f$ as precondition;

- $add_f \subseteq \mathcal{A}$, $\forall f \in \mathcal{F}$, set of actions that have fluent $f$ as add effect;

- $del_f \subseteq \mathcal{A}$, $\forall f \in \mathcal{F}$, set of actions that have fluent $f$ as delete effect;

Variables are defined for each layer $1 \leq t \leq T$ in the planning graph. There are variables for the actions and there are variables for the possible state changes a fluent can make, but only those variables that are reachable and relevant by planning graph analysis are instantiated. For all $a \in \mathcal{A}$, $t \in 1, ..., T$ we have the action variables

$$y_{a,t} = \left\{ \begin{array}{ll} 1 & \text{if action } a \text{ is executed in period } t, \\ 0 & \text{otherwise.} \end{array} \right.$$

The "no-op" actions are not included in the $y_{a,t}$ variables but are represented separately by the state change variable $x_{f,t}^{maintain}$.

Optiplan is based on the state change formulation (Vossen et al., 1999). In this formulation fluents are not represented explicitly, instead state change variables are used to model transitions in the world state. That is, a fluent is true if and only if it is added to the state by $x_{f,t}^{add}$ or $x_{f,t}^{preadd}$, or if it is persisted from the previous state by $x_{f,t}^{maintain}$. Optiplan extends the state change formulation (Vossen et al., 1999) by introducing an extra state change variable, $x_{f,t}^{del}$, that allows actions to delete fluents without requiring them as preconditions. The original state change formulation did not allow for these actions, so therefore we added these new state change variables to keep track of such state changes and altered the model to take these new variables into account. In the IPC4 domains of Airport and PSR there are many actions that delete fluents without requiring them as preconditions, therefore making the original state change formulation ineffective. Also, Optiplan instantiates only those variables and constraints that are reachable and relevant through planning graph analysis, and therefore creates a smaller encoding than the original one. For all $f \in \mathcal{F}$, $t \in 1, ..., T$ we have the following state change variables:

$$x_{f,t}^{maintain} = \left\{ \begin{array}{ll} 1 & \text{if fluent } f \text{ is propagated in period } t, \\ 0 & \text{otherwise.} \end{array} \right.$$

$$x_{f,t}^{preadd} = \left\{ \begin{array}{ll} 1 & \text{if action } a \text{ is executed in period } t \text{ such that } a \in pre_f \wedge a \notin del_f, \\ 0 & \text{otherwise.} \end{array} \right.$$

$$x_{f,t}^{predel} = \left\{ \begin{array}{ll} 1 & \text{if action } a \text{ is executed in period } t \text{ such that } a \in pre_f \wedge a \in del_f, \\ 0 & \text{otherwise.} \end{array} \right.$$

$$x_{f,t}^{add} = \left\{ \begin{array}{ll} 1 & \text{if action } a \text{ is executed in period } t \text{ such that } a \notin pre_f \wedge a \in add_f, \\ 0 & \text{otherwise.} \end{array} \right.$$

$$x_{f,t}^{del} = \left\{ \begin{array}{ll} 1 & \text{if action } a \text{ is executed in period } t \text{such that } a \notin pre_f \wedge a \in del_f, \\ 0 & \text{otherwise.} \end{array} \right.$$

In summary: $x_{f,t}^{maintain} = 1$ if the truth value of a fluent is propagated; $x_{f,t}^{preadd} = 1$ if an action is executed that requires a fluent and does not delete it; $x_{f,t}^{predel} = 1$ if an action is executed that requires a fluent and deletes it; $x_{f,t}^{add} = 1$ if an action is executed that does not require a fluent and adds it; and $x_{f,t}^{del} = 1$ if an action is executed that does not require a fluent and deletes it. The complete IP formulation of Optiplan is given by the following objective function and constraints.

## 4.1 Objective

For classical AI planning problems, no minimization or maximization is required, instead we want to find a feasible solution. The search for a solution, however, may be guided by an objective function such as the minimization of the number of actions, which is currently implemented in Optiplan. The objective function is given by:

$$\min \quad \sum_{a \in \mathcal{A}} \sum_{i \in T} y_{a,t} \tag{1}$$

Since the constraints guarantee feasibility we could have used any linear objective function. For example, we could easily set up an objective to deal with cost-sensitive plans (in the context of non-uniform action cost), utility-sensitive plans (in the context of over-subscription and goal utilities), or any other metric that can be transformed to a linear expression. Indeed this flexibility to handle any linear objective function is one of the advantages of IP formulations.

## 4.2 Constraints

The requirements on the initial and goal transition are given by:

$$x_{f,0}^{add} = 1 \qquad \forall f \in \mathcal{I} \tag{2}$$

$$x_{f,0}^{add}, x_{f,0}^{maintain}, x_{f,0}^{preadd} = 0 \qquad \forall f \notin \mathcal{I} \tag{3}$$

$$x_{f,T}^{add} + x_{f,T}^{maintain} + x_{f,T}^{preadd} \geq 1 \qquad \forall f \in \mathcal{G} \tag{4}$$

Where constraints (2), and (3) add the initial fluents in step 0 so that they can be used by the actions that appear in the first layer (step 1) of the planning graph. Constraints (4) represent the goal state requirements, that is, fluents that appear in the goal must be added or propagated in step $T$.

The state change variables are linked to the actions by the following effect implication constraints. For each $f \in \mathcal{F}$ and $1 \leq t \leq T$ we have:

$$\sum_{a \in add_f \setminus pre_f} y_{a,t} \geq x_{f,t}^{add} \tag{5}$$

$$y_{a,t} \leq x_{f,t}^{add} \qquad \forall a \in add_f \setminus pre_f \tag{6}$$

$$\sum_{a \in del_f \setminus pre_f} y_{a,t} \geq x_{f,t}^{del} \tag{7}$$

$$y_{a,t} \leq x_{f,t}^{del} \qquad \forall a \in del_f \setminus pre_f \tag{8}$$

$$\sum_{a \in pre_f \setminus del_f} y_{a,t} \geq x_{f,t}^{preadd} \tag{9}$$

$$y_{a,t} \leq x_{f,t}^{preadd} \qquad \forall a \in pre_f \setminus del_f \tag{10}$$

$$\sum_{a \in pre_f \wedge del_f} y_{a,t} = x_{f,t}^{predel} \tag{11}$$

Where constraints (5) to (11) represent the logical relations between the action and state change variables. The equality sign in (11) is because all actions that have $f$ as a precondition and as a delete effect are mutually exclusive. This also means that we can substitute out the $x_{f,t}^{predel}$ variables, which is what we have done in the implementation of Optiplan. We will, however, use the variables here for clarity. Mutexes also appear between different state change variables and these are expressed by constraints as follows:

$$x_{f,t}^{add} + x_{f,t}^{maintain} + x_{f,t}^{del} + x_{f,t}^{predel} \leq 1 \tag{12}$$

$$x_{f,t}^{preadd} + x_{f,t}^{maintain} + x_{f,t}^{del} + x_{f,t}^{predel} \leq 1 \tag{13}$$

Where constraints (12) and (13) restrict certain state changes from occurring in parallel. For example, $x_{f,t}^{maintain}$ (propagating fluent $f$ at step $t$) is mutually exclusive with $x_{f,t}^{add}$, $x_{f,t}^{del}$, and $x_{f,t}^{predel}$ (adding or deleting $f$ at $t$).

Finally, the backward chaining requirements and binary constraints are represented by:

$$x_{f,t}^{preadd} + x_{f,t}^{maintain} + x_{f,t}^{predel} \leq x_{f,t-1}^{preadd} + x_{f,t-1}^{add} + x_{f,t-1}^{maintain} \qquad \forall f \in \mathcal{F}, t \in 1, ..., T \tag{14}$$

$$x_{f,t}^{preadd}, x_{f,t}^{predel}, x_{f,t}^{add}, x_{f,t}^{del}, x_{f,t}^{maintain} \in \{0,1\} \tag{15}$$

$$y_{a,t} \in \{0,1\} \tag{16}$$

Where constraints (14) describe the backward chaining requirements, that is, if a fluent $f$ is added or maintained in step $t-1$ then the state of $f$ can be changed by an action in step $t$ through $x_{f,t}^{preadd}$, or $x_{f,t}^{predel}$, or it can be propagated through $x_{f,t}^{maintain}$. Constraints (15) and (16) are the binary constraints for the state change and action variables respectively.

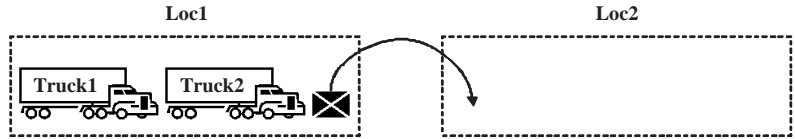

Figure 1: A simple logistics example

## 4.3 Example

In this example, we show how some of the constraints are initialized and we comment on the interaction between the state change variables and the action variables.

Consider a simple logistics example in which there are two locations, two trucks, and one package. The package can only be transported from one location to another by one of the trucks. We built a formulation for three plan steps. The initial state is that the package and the trucks are all at location 1 as given in Figure 1. The initial state constraints are:

$$x^{add}_{\textbf{pack1\_at\_loc1}, 0} = 1$$
$$x^{add}_{\textbf{truck1\_at\_loc1}, 0} = 1$$
$$x^{add}_{\textbf{truck2\_at\_loc1}, 0} = 1$$
$$x^{add}_{f, 0}, x^{maintain}_{f, 0}, x^{preadd}_{f, 0} = 0 \qquad f \neq \mathcal{I}$$

The goal is to get the package at location 2 in three plan steps, which is expressed as follows:

$$x^{add}_{\textbf{pack1\_at\_loc2}, 3} + x^{maintain}_{\textbf{pack1\_at\_loc2}, 3} + x^{preadd}_{\textbf{pack1\_at\_loc2}, 3} \geq 1$$

We will not write out all effect implication constraints, but we will comment on a few of them. If $x^{add}_{f, t} = 1$ for a certain fluent $f$, then we have to execute at least one action $a$ that has $f$ as an add effect and not as a precondition. For example:

$$y_{\textbf{unload\_truck1\_at\_loc2}, t} + y_{\textbf{unload\_truck2\_at\_loc2}, t} \geq x^{add}_{\textbf{pack1\_at\_loc2}, t}$$

The state changes for $x^{del}_{f, t}$ and $x^{preadd}_{f, t}$ have a similar requirement, that is if we change the state through *del* or *preadd* then we must execute at least one action $a$ with the corresponding effects. On the other hand, if we execute an action $a$ then we must change all fluent states according to the effects of $a$. For example:

$$y_{\textbf{unload\_truck1\_at\_loc2}, t} \leq x^{add}_{\textbf{pack1\_at\_loc2}, t}$$
$$y_{\textbf{unload\_truck1\_at\_loc2}, t} \leq x^{preadd}_{\textbf{truck\_at\_loc2}, t}$$
$$y_{\textbf{unload\_truck1\_at\_loc2}, t} = x^{predel}_{\textbf{pack1\_in\_truck1}, t}$$

There is a one-to-one correspondence (note the equality sign) between the execution of actions and the $x^{predel}_{f, t}$ state change variables. This is because, actions that have the same *predel* effect must be mutex. Mutexes are also present between state changes. For example, a fluent $f$ that is maintained (propagated) cannot be added or deleted. The only two state changes that are not mutex are the *add* and the *preadd*. This is because the *add* state change behaves like the *preadd* state change if the corresponding fluent is already present in the state of the world. This is why we introduce two separate mutex constraints, one that includes the *add* state change and one that includes the *preadd*. An example for the constraints on the mutex state changes are as follows:

$$x^{add}_{\textbf{pack1\_in\_truck1}, t} + x^{maintain}_{\textbf{pack1\_in\_truck1}, t} + x^{del}_{\textbf{pack1\_in\_truck1}, t} + x^{predel}_{\textbf{pack1\_in\_truck1}, t} \leq 1$$
$$x^{preadd}_{\textbf{pack1\_in\_truck1}, t} + x^{maintain}_{\textbf{pack1\_in\_truck1}, t} + x^{del}_{\textbf{pack1\_in\_truck1}, t} + x^{predel}_{\textbf{pack1\_in\_truck1}, t} \leq 1$$

The state of a fluent can change into another state only if correct state changes have occurred previously. Hence, a fluent can be deleted, propagated, or used as preconditions in step $t$ if and only if it was added or propagated in step $t - 1$. For example:

$$x^{preadd}_{\textbf{pack1\_in\_truck1}, t} + x^{maintain}_{\textbf{pack1\_in\_truck1}, t} + x^{predel}_{\textbf{pack1\_in\_truck1}, t} \leq$$
$$x^{preadd}_{\textbf{pack1\_in\_truck1}, t-1} + x^{add}_{\textbf{pack1\_in\_truck1}, t-1} + x^{maintain}_{\textbf{pack1\_in\_truck1}, t-1}$$

| $t = 0$ | $t = 1$ | $t = 2$ | $t = 3$ |
|---|---|---|---|
| $x^{add}_{\textbf{pack1\_at\_loc1},0}$ | $y_{\textbf{load\_truck1\_at\_loc1}},1$ $x^{add}_{\textbf{pack1\_in\_truck1},1}$ $x^{predel}_{\textbf{pack1\_at\_loc1},1}$ | $y_{\textbf{drive\_truck1\_loc1\_loc2}},2$ $x^{maintain}_{\textbf{pack1\_in\_truck1},2}$ | $y_{\textbf{unload\_truck1\_at\_loc2}},3$ $x^{add}_{\textbf{pack1\_at\_loc2},3}$ $x^{predel}_{\textbf{pack1\_in\_truck1},3}$ |
| $x^{add}_{\textbf{truck1\_at\_loc1},0}$ | $x^{preadd}_{\textbf{truck1\_at\_loc1},1}$ | $x^{add}_{\textbf{truck1\_at\_loc2},2}$ $x^{predel}_{\textbf{truck1\_at\_loc1},2}$ | $x^{preadd}_{\textbf{truck1\_at\_loc2},3}$ |
| $x^{add}_{\textbf{truck2\_at\_loc1},0}$ | $x^{maintain}_{\textbf{truck2\_at\_loc1},1}$ | | |

Table 1: Solution to the simple logistics example. All displayed variables have value 1 and all other variables have value 0.

This simple problem has a total of 107 variables (41 action and 66 state change) and 91 constraints. However, planning graph analysis fixes 53 variables (28 action and 25 state change) to zero. After substituting these values and applying presolve techniques that are built in the ILOG CPLEX solver, this problem has only 13 variables and 17 constraints. The solution for this example is given in Table 1. Note that, when there are no actions that actively delete $f$, there is nothing that ensures $x^{maintain}_{f,t}$ to be true whenever $f$ was true in the preceding state (for example, see the fluent **truck2\_at\_loc1**). Since negative preconditions are not allowed, having the option of letting $x^{maintain}_{f,t}$ be false when it should have been true cannot cause actions to become executable when they should not be. We will not miss any solutions because constraints (4) ensure that the goal fluents are satisfied, therefore forcing $x^{maintain}_{f,t}$ to be true whenever this helps us generate a plan.

## 5. Experimental Results

First we compare Optiplan with the original state change model, and then we check how Optiplan performed in the IPC of 2004.

Optiplan and the original state change formulation are implemented in two different languages. Optiplan is implemented in C++ using Concert Technology, which is a set of libraries that allow you to embed ILOG CPLEX optimizers (ILOG Inc., 2002), and the original state change model is implemented in AMPL (Fourer, Gay, & Kernighan, 1993), which is a modeling language for mathematical programming. In order to compare the formulations that are produced by these two implementations, they are written to an output file using the MPS format. MPS is a standard data format that is often used for transferring linear and integer linear programming problems between different applications. Once the MPS file, which contains the IP formulation for the planning problem, is written, it is read and solved by ILOG CPLEX 8.1 on a Pentium 2.67 GHz with 1.00 GB of RAM.

Table 3 shows the encoding size of the two implementations, where the encoding size is characterized by the number of variables and the number of constraints in the formulation. Both the encoding size before and after applying ILOG CPLEX presolve is given. Presolve is a problem reduction technique (Brearley, Mitra, & Williams, 1975) that helps most linear programming problems by simplifying, reducing and eliminating redundancies. In short, presolve tries to remove redundant constraints and fixed variables from the formulation, and aggregate (substitute out) variables if possible.

| Problem | State change model | | | | Optiplan | | | |
|---|---|---|---|---|---|---|---|---|
| | Before presolve | | After presolve | | Before presolve | | After presolve | |
| | #Var. | #Cons. | #Var. | #Cons. | #Var. | #Cons. | #Var. | #Cons. |
| bw-sussman | 486 | 878 | 196 | 347 | 407 | 593 | 105 | 143 |
| bw-12step | 3900 | 7372 | 1663 | 3105 | 3534 | 4998 | 868 | 1025 |
| bw-large-a | 6084 | 11628 | 2645 | 5022 | 5639 | 8690 | 1800 | 2096 |
| att-log0 | 1932 | 3175 | 25 | 35 | 117 | 149 | 0 | 0 |
| log-easy | 24921 | 41457 | 1348 | 2168 | 2534 | 3029 | 437 | 592 |
| log-a | 50259 | 85324 | 3654 | 6168 | 5746 | 7480 | 1479 | 2313 |

Table 2: Encoding size of the original state change formulation and Optiplan before and after ILOG CPLEX presolve. #Var. and #Cons. give the number of variables and constraints respectively.

| Problem | State change model | | | | Optiplan | | | |
|---|---|---|---|---|---|---|---|---|
| | #Var. | #Cons. | #Nodes | Time | #Var. | #Cons. | #Nodes | Time |
| bw-sussman | 196 | 347 | 0 | 0.01 | 105 | 143 | 0 | 0.01 |
| bw-12step | 1663 | 3105 | 19 | 4.28 | 868 | 1025 | 37 | 1.65 |
| bw-large-a | 2645 | 5022 | 2 | 8.45 | 1800 | 2096 | 0 | 0.72 |
| bw-large-b | 6331 | 12053 | 14 | 581.92 | 4780 | 5454 | 10 | 72.58 |
| att-log0 | 25 | 35 | 0 | 0.01 | 0 | 0 | 0 | 0.01 |
| att-log1 | 114 | 164 | 0 | 0.03 | 29 | 35 | 0 | 0.01 |
| att-log2 | 249 | 371 | 10 | 0.07 | 81 | 99 | 0 | 0.01 |
| att-log3 | 2151 | 3686 | 15 | 0.64 | 181 | 228 | 0 | 0.03 |
| att-log4 | 2147 | 3676 | 12 | 0.71 | 360 | 507 | 0 | 0.04 |
| att-loga | 2915 | 4968 | 975 | 173.56 | 1479 | 2312 | 19 | 2.71 |
| rocket-a | 1532 | 2653 | 517 | 32.44 | 991 | 1644 | 78 | 5.48 |
| rocket-b | 1610 | 2787 | 191 | 9.90 | 1071 | 1788 | 24 | 3.12 |
| log-easy | 1348 | 2168 | 43 | 0.96 | 437 | 592 | 0 | 0.04 |
| log-a | 3654 | 6168 | 600 | 145.31 | 1479 | 2313 | 19 | 2.66 |
| log-b | 4255 | 6989 | 325 | 96.47 | 1718 | 2620 | 187 | 14.06 |
| log-c | 5457 | 9111 | 970 | 771.36 | 2413 | 3784 | 37 | 16.07 |

Table 3: Performance and encoding size of the original state change formulation and Optiplan. #Var. and #Cons. give the number of variables and constraints after ILOG CPLEX presolve, and #Nodes give the number of nodes explored during branch-and-bound before finding the first feasible solution.

From the encoding size before presolve, which is the actual encoding size of the problem, we can see how significant the use of planning graphs is. Optiplan, which instantiates only those fluents and actions that are reachable and relevant through planning graph analysis, produces encodings that in some cases are over one order of magnitude smaller than the encodings produced by the original state change model, which instantiates all ground fluents and actions. Although the difference in the encoding size reduces substantially after applying presolve, planning graph analysis still finds redundancies that presolve fails to detect. Consequently, the encodings produced by Optiplan are still smaller than the encodings that are produced by the original state change model.

The performance (and the encoding size after presolve) of Optiplan and the original state change model are given in Table 3. Performance is measured by the time to find the first feasible solution. The results show the overall effectiveness of using planning graph analysis. For all problems Optiplan not only generates smaller encodings it also performs better than the encodings generated by the state change model.

## 5.1 IPC Results

Optiplan participated in the propositional domains of the optimal track in the IPC 2004. In this track, planners could either minimize the number of actions, like BFHSP and Semsyn; minimize makespan, like CPT, HSP*a, Optiplan, Satplan04, and TP-4; or minimize some other metric.

The IPC results of the makespan optimal planners are given in Figure 2. All results were evaluated by the competition organizers by looking at the runtime and plan quality graphs. Also, all planners were compared to each other by estimating their asymptotic runtime and by analyzing their solution quality performance. Out of the seven competition domains, Optiplan was judged second best in four of them. This is quite remarkable because integer programming has hitherto not been considered competitive in planning.

Optiplan reached second place in the Optical Telegraph and the Philosopher domains. In these domains Optiplan is about one order of magnitude slower than Satplan04, but it clearly outperforms all other participating planners. In the Pipesworld Tankage domain, Optiplan was awarded second place together with Satplan04, and in the Satellite domain Optiplan, CPT, and Semsyn all tied for second place. In the other domains Optiplan did not perform too well. In the Airport domain, Optiplan solves the first 17 problems and problem 19, but it takes the most time to do so. For the Pipesworld Notankage and the PSR domains, Optiplan not only is the slowest it also solves the fewest number of problems among the participating planners.

In looking at the domains and problems where Optiplan has difficulty scaling, we notice that these are problems that lead to very large IP encodings. Since the size of the encoding is a function of plan length, Optiplan often fails to solve problems that have long solution plans. One way to resolve this issue is to de-link the encoding size from solution length, which is what we have done in some of our recent work (van den Briel, Vossen, & Kambhampati, 2005). In fact, in the year following the IPC4 we developed novel IP encodings that (1) model transitions in the individual fluents as separate but loosely coupled network flow problems, and that (2) control the encoding length by generalizing the notion of parallelism.

## 6. Conclusions

The Optiplan planning system performs significantly better than the original state change model by Vossen and his colleagues (1999). It performed respectably at the IPC4, but still lags behind SAT- and CSP-based planners, like Blackbox(Chaff), Satplan04(Siege), and GP-CSP. We believe, however, that this performance gap is not because IP techniques are inferior to SAT and CSP, but rather a reflection of the types of IP formulations that have been tried so far. Specifically, the encodings that have been tried until now have not been tailored to the strengths of the IP solvers (Chandru & Hooker, 1999).

Our experience with Optiplan has encouraged us to continue working on improved IP formulations for AI planning. In our recent work (van den Briel, Vossen, & Kambhampati, 2005) we model fluents as loosely coupled network flow problems and control the encoding length by generalizing the notion of parallelism. The resulting IP encodings are solved within a branch-and-cut algorithm and yield impressive results. Also, this new approach has been shown to be highly competitive with the state-of-the-art SAT-based planners.

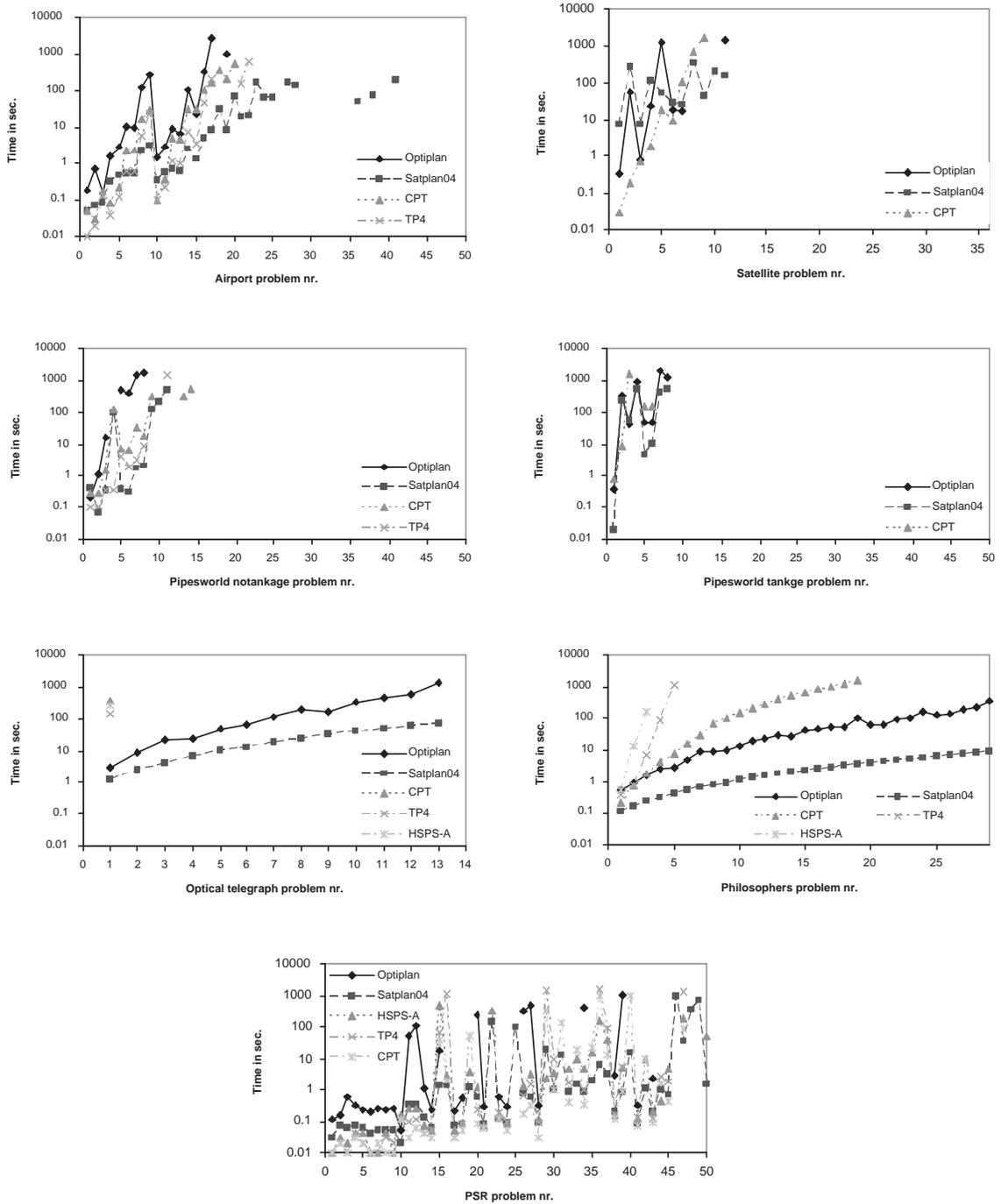

Figure 2: IPC 2004 results for the makespan optimal planners.